# A Scenario-Based Functional Testing Approach to Improving DNN Performance

Hong Zhu, Thi Minh Tam Tran, Aduen Benjumea and Andrew Bradley
School of Engineering, Computing and Mathematics, Oxford Brookes University
Oxford OX33 1HX, UK; Email: hzhu@brookes.ac.uk

## Abstract

This paper proposes a scenario-based functional testing approach for enhancing the performance of machine learning (ML) applications. The proposed method is an iterative process that starts with testing the ML model on various scenarios to identify areas of weakness. It follows by a further testing on the suspected weak scenarios and statistically evaluate the model's performance on the scenarios to confirm the diagnosis. Once the diagnosis of weak scenarios is confirmed by test results, the treatment of the model is performed by retraining the model using a transfer learning technique with the original model as the base and applying a set of training data specifically targeting the treated scenarios plus a subset of training data selected at random from the original train dataset to prevent the so-call catastrophic forgetting effect. Finally, after the treatment, the model is assessed and evaluated again by testing on the treated scenarios as well as other scenarios to check if the treatment is effective and no side-effect caused. The paper reports a case study with a real ML deep neural network (DNN) model, which is the perception system of an autonomous racing car. It is demonstrated that the method is effective in the sense that DNN model's performance can be improved. It provides an efficient method of enhancing ML model's performance with much less human and compute resource than retrain from scratch.

## 1 Introduction

In recent years, machine learning (ML), especially deep neural networks (DNNs), has been increasingly employed by critical applications such as security protection, autonomous vehicles, communication, medical research etc. In order to understand the performance and limitations of the system, it is necessary to perform extensive testing covering the complete variety of possible scenarios that the system may encounter. The basic idea of scenario-based testing (SBT) is to split test cases into a number of subsets so that each subset tests the system on one possible operation scenario and the performance of the system is assessed and evaluated on one scenario at a time. It has been widely used in traditional software engineering and proven to be efficient and effective. However, its application to testing DNN is still very limited because of DNN's fundamental differences from the traditional software. The only exception is perhaps in the testing of autonomous vehicles (AV), where SBT is required by ISO 26262 standard for road vehicle functional safety [1] and pursued as the state of art in the AV testing community [2]. However, the research on SBT for AV is limited to the validation, verification and safety assessment of AV rather than quality improvement [3]. A problem that remains open is how to improve ML model's performance based on the results of testing.

Traditionally, faults in program detected by testing can be fixed through debugging (i.e., by modifying the program code), and therefore improving the quality of the software under test. However, ML models like DNN cannot be debugged through manually modifying the weights of the links between neurons and adding/removing neurons in the network [4]. To improve the performance of a neural network, one has to retrain the model with additional training data. Typically, the erroneous behaviours of a DNN once detected can be fixed by adding the error-inducing inputs to the training data set and/or possibly by changing the model structure and hyperparameters. However, such error-inducing test cases are rare and insufficient to improve the model's performance. It can be a labour intensive and expensive process to acquire and label such data. As Tian *et al* pointed out [5], how to use data obtained from testing to improve ML models is still a challenging problem, which has been noted by large IT companies like Google and Tesla [6, 7].

In this paper we address this problem at two levels of abstraction. At the methodology level, we propose an exploratory iterative process model of SBT, which includes statistical scenarios analysis for diagnosing ML model's weakness and treatment of the ML model targeting the diagnosed weak scenarios in order to improve its quality. At the technology level, we propose the uses of a combination of ML methods and techniques to treat ML models on weak scenarios, which include employing transfer learning techniques and data augmentation.

The paper is organised as follows. Section 2 reviews related work. Section 3 presents the proposed SBT method. Section 4 reports a case study to demonstrate the effectiveness and efficiency of the proposed method. It is a real ML model of the perception system of an autonomous racing car. Section 5 concludes the paper with a discussion of future work.

## 2 Related Work

Testing ML applications has been an active research topic in recent years. This section briefly reviews the current state of art with focus on functional testing. The research on testing other quality attributes of ML applications, such as robustness and fairness, etc., are omitted.

### 2.1 Testing Process Models

The proposed methodology is partly inspired by the use case driven methodology of Object-Oriented (OO) software development [8], in which SBT plays the central role. The notion of scenario is defined as a linear sequence of interactions (i.e., input/output) between the system and its user as an instance of a use case. SBT is performed manually by designing test cases, generating test data and checking system's output against the excepted outputs according to the specifications of scenarios. Scenarios are usually informally or semi-formally represented, for example, in use case diagrams, activity diagrams and/or state machines in UML. The quality of the software under test are improved by debugging program code according to the bug reports. It has been proven to be an effective and efficient software quality assurance methodology. Unfortunately, this cannot be straightforwardly applied to the development of ML based computer applications because ML models cannot be debugged by editing the model parameters.

SBT has been pursued as the state of art in the AV testing community. The ISO standard 26262 for road vehicle functional safety [1] evolved from its early versions that are based on the best practice of traditional software engineering. The current version inherited from its earlier versions to include scenario as one of the key concepts in its framework for AV safety assurance. In particular, SBT is required by its waterfall (or V) model of functional safety engineering [1, 9-11]. In the AV testing community, it is widely accepted that the notion of scenario can be defined as a temporal sequence of scenes while a scene is a snapshot of the environment of the AV. For a scenario to be used as a test case, it is also associated with the expected behaviour of the system. In [12], Menzel, Bagschik and Maurer analysed the notion of scenarios in ISO standard 26262 and the requirements on scenario representations at different stages of development process. They distinguished the notion of scenarios at three different levels of abstraction:

- *functional scenarios*, which are described in natural language by domain experts at the requirements stage,
- *logic scenarios*, which are formally specified by a set of state variables and their value ranges, and
- *concrete scenarios*, in which each state variable is assigned a specific value in the corresponding range of the logic scenario. They are the test data for executing tests of the scenario.

In the waterfall development process, at requirements stage, the functional scenarios are elicited and hazardous scenarios are identified. From functional scenarios, logical scenarios are derived and finally transformed or converted into concrete scenarios for the test executions. The distinction between these notions of scenarios is now widely adopted by the AV testing community. Based on this, in [11], Neurohr *et al* analysed the considerations around scenario-based testing at each stage of a waterfall framework and reviewed how these considerations are addressed in the literature.

In recent years, much work has been reported in the literature to derive scenarios for testing AV manually and generate them automatically [13-19]. All of these approaches proposed and studied in the literature belong to confirmatory testing methods. That is, they apply scenario-based testing to validate and verify the system's conformation to requirements, such as meeting the functional safety requirements.

In [20], based on datamorphic testing methodology, Zhu *et al* proposed a more general waterfall model of SBT of ML applications. The notion of scenario is defined as *operation conditions* of the system, which is more general than the notion in AV testing because many ML applications are not interactive. Its process model consists of three stages. At the first state, *scenario analysis*, the normal scenarios as well as abnormal scenarios are identified and defined. Where normal scenarios are the operation conditions that occurs most frequently, while abnormal scenarios rarely occur, such as hazardous situations. Thus, adequate test data can be obtained relatively easier for normal scenarios, while more difficult for abnormal scenarios. At the second stage, *realisation*, a test system is designed and implemented to actually build test datasets for all scenarios and to perform testing and analyse test results. Typically, test data for normal scenarios are collected, while for abnormal scenarios, test datasets are generated via implementation and application of datamorphisms (i.e., data augmentations) that transform test data for normal scenarios to those for abnormal scenarios. The final stage is *test execution* in which the ML model is actually tested on all scenarios and the performance of the model is evaluated.

In contrast to the above SBT methods, the method proposed in this paper belongs to exploratory testing, which aims at discovering unknow problems of the system under test. The idea is to organise the whole process as iterative cycles of exploring the ML model under test to discover the scenarios in which its performances are unsatisfactory and then improving the performances on such scenarios. Although our process is in analogue to the cycles of testing and debugging for traditional software, it is widely recognised as a challenge for how this can be done [5].

## 2.2 Test Oracle Problem

One of the most challenging problems for functional testing of ML models is the so-called *test oracles problem*, i.e. how to check the correctness of the output from a ML model on test cases. A solution that has been pursued by many researchers in recent years is metamorphic testing, where a metamorphic relation is an assertion about software's behaviour on multiple interrelated test cases. They are used to check software correctness as well as to generate test cases. However, how to find metamorphic relations has been a hard problem. The datamorphic approach to deriving metamorphic relations was proposed in [21] and further developed in [22, 23]. The basic idea is to first identify operation scenarios of the system, then develop semantic preserving or semantic transforming datamorphisms to transform test data between the scenarios. The metamorphisms can then be easily derived from the datamorphisms.

## 2.3 Test Data Generation

Another challenging problem of functional testing is how to generate test data. For most machine learning applications, a large volume of real data is available for the common operation scenarios. However, real data can be difficult and expensive to obtain on adverse operation conditions, such as in hazardous scenarios. For example, for testing AV, sufficient test data on good weather and traffic conditions can be obtained by recording real world uses of vehicles. However, such recorded data are far from sufficient for rare weather and traffic situations or in dangerous scenarios involving accidents. A solution to this problem explored by researchers and reported in the literature is to employ data augmentations to generate synthetic test data from real data. For example, Tian et al's *DeepTest* system [5] employed nine different image transformations to augment data for testing AV. Their augmentations were changing brightness, changing contrast, translation, scaling, horizontal shearing, rotation, blurring, adding fog effect, and adding rain effect. They used photoshop functions to implement the augmentations of adding fog and rain effects, while other augmentations were implemented through simple image processing algorithms. Hasirlioglu and Riener [25] developed a digital augmentation algorithm based on a theoretical model of the effect of rain on images and sensors. The technique is extended by Musat et al. [41] to simulate the combinations of multiple weather conditions.

The uses of generative adversarial networks have also been explored to generate test cases. Among the earliest attempts are Zhang *et al.* and Zhu *et al.*'s work [20, 26]. Zhang *et al.* trained a GAN to generate test data for snowy and rainy scenarios from sunny conditions. Zhu *et al.* also employed a generative adversarial network (GAN) AttGan [20] to change the features of face images in testing face recognition ML models. They conducted experiments to compare the test results using synthetic images against the test results of using real images. They concluded that such synthetic images are valid test cases that produced test results consistent with those using real images.

### 2.4 Test Adequacy and Coverage

Requirements coverage and scenario coverage are the most widely used adequacy criteria in testing AV [2], where requirements coverage means all functional requirements are tested, while scenario coverage means all scenarios identified in safety requirements analysis are tested. When test cases are generated by applying data augmentations that each represents a different operation scenario, more complicated scenario coverage criteria can be defined. Zhu *et al.* [20] proposed a set of such adequacy criteria for coverage mutant test cases, where a seed test case is the original test data, while a mutant test case is the test data generated by applying a data augmentation operator. The *First Order Mutant Coverage* requires a test set to contain all seed test cases and the first order mutant test cases. The *Second Order Mutant Coverage* requires to contain all seed test cases, 1st and 2nd order mutants. In general, a $K$'th *Order Mutant Coverage* requires that a test set contains all $n$'th order mutants for all $K > n > 0$, where $K > 1$. The *Mutation Combination Complete* criterion requires the test set contain all combinations of high order mutants, which is equivalent to the exhaustive test if the datamorphisms satisfy certain algebraic laws. They also devised algorithms to generate test datasets that meets these adequacy criteria. Considering the combinations of augmentations to be too expensive because a large number of test cases can be generated, Tian *et al.* [5] proposed an algorithm that searches for those combinations that can increase neuron coverage best.

A number of test adequacy criteria have also been proposed and studied in the literature inspired by structural testing of traditional programs. Among the most well-known are Pei *et al.*'s neuron coverage [24] inspired by statement coverage, Ma *et al.*'s neuron output range coverage [27], Ma *et al.*'s neuron output combination coverage inspired by combinatorial testing [28], Sun *et al.* decision influence coverage inspired by the MC/DC adequacy criterion [29, 30], and Xie *et al.*'s neuron path coverage criteria [31] inspired by control flow and data flow testing methods.

In [32], Kim *et al.* proposed surprise-based approach to structural testing of neural networks, which requires test cases to be spread far from each other and far away from the training data where the distances are measured by the distances between the neuron activation traces of the test data and the training data.

These adequacy criteria have been empirically evaluated on whether the coverage metric is a reliable predictor of the correctness of the ML model under test, whether the coverage metric is correlated to the output impartiality, whether the calculation of the coverage metric is efficient and scalable, and whether using the coverage criterion as the guide to generate test cases can help to detect error of the ML model, etc. However, Harel-Canada *et al.* conducted experiments with neuron coverage criterion and demonstrated that such test cases are not natural. They questioned if neuron coverage is meaningful in testing [33]. The same question remains for all other structural test adequacy criteria.

## 3 The Proposed Methodology

This section presents the proposed methodology for scenario-based functional testing of ML models.

The proposed testing process is an iterative cycle of the following steps.

- *Assumption*: identifying scenarios in which system's performance needs improvement. This means testing the ML model and evaluating its performances. The goal is to identify the scenarios in which the model's performance is unsatisfactory. Typically, weak scenarios are identified through manual inspections of test cases on which the system fails. Assumptions on system's weakness are formed and represented as scenarios, which are called *suspected* weak scenarios.
- *Diagnosis*: confirming the assumptions via further testing ML model on the suspected scenarios and statistically analyse ML model's performance on these scenarios so that the assumption can be confirmed based on test results. Once confirmed, such a scenario is called a *diagnosed* weak scenario.
- *Treatment*: retraining the ML model with additional training data that target the diagnosed weak scenarios. This will involve either collecting additional training data or generating data from existing ones by applying data augmentations. Attention should be paid to prevent side-effects of the treatment, such as the *forget* phenomenon that the performance on other scenarios, even the overall performance on whole input space, decreases. Such scenarios are called *treated* scenarios of the result model.
- *Evaluation*: testing the treated ML model after

retraining and evaluating the effect of the treatment. This means testing the model on the treated scenarios as well as other scenarios. The evaluation should aim at two goals: (a) checking the effectiveness of treatment, i.e., whether the result model actually improves performance on treated scenarios; (b) checking if there is any side-effect, i.e., if the result model improves overall performance and remain in good performances on other scenarios. The failure test cases will feed into the next cycle of the process.

The key technical problem of the above process is how to treat a ML model on a specific weak scenario without causing side-effects. Our proposed solution is:

- *Transfer learning*: to apply a transfer learning technique, i.e., to use the existing model as the base for retraining.
- *Targeted training*: to use a set of training data that represent the scenarios to be treated.
- *Prevention of side-effect*: to include a subset of the original training data in the retraining dataset to prevent the so-called *forget* effect.

# 4 Case Study

To demonstrate the applicability of the proposed method, this section reports a case study with the perception system of an autonomous racing car.

## 4.1 Background

94% of all road traffic accidents are caused by driver error [34], and thus replacing the driver with autonomous control offers the potential for a significant improvement in road safety. However, safe deployment of AV on the road requires significant testing in a controlled environment to ensure that they will operate in a reliable, robust, and safe manner.

The advent of autonomous racing provides an ideal test-bed for testing and developing AV technologies. The fierce competition of motorsport demands the highest levels of accuracy from every element of the system while running at the high speeds involved in racing. Small performance margins (e.g., due to changing weather conditions) can make the difference between winning and losing a race - thus requiring detailed analysis of the performance of every subsystem within the vehicle in every possible condition. The closed nature of the racetrack provides the perfect 'sandboxed' environment, where errors or failures within development of the hardware or software systems do not result in catastrophic consequences [35].

Whilst the sandboxed testing environment provides a *safe* place to test AV systems, there remain challenges to testing the vehicle. Real-world testing is expensive, time-consuming, and the weather cannot be scheduled. Thus, the ability to virtually test the AV control system components in a variety of conditions provides significant efficiency potential.

The majority of AV and autonomous racing control systems operate using a pipeline of subsystems - typically comprising perception, decision making, path planning and control [36-38]. Since the perception system is the first subsystem in the pipeline, it is critical to ensure the highest level of performance to mitigate the propagation of errors through the pipeline.

This case study focuses upon the use of the scenario based functional testing methodology to test and develop the AI perception subsystem of an autonomous racing vehicle, thereby identifying potential areas of weakness. This enables the development of strategic, targeted improvements to the AI component, thereby improving performance of the whole system in a highly efficient manner.

## 4.2 Formula SAE Competition

Numerous autonomous vehicle competitions exist - ranging from small-scale racing cars like *FormulaPi*[1] and *F1Tenth*[2] to controlling fleets of self-navigating, lane-changing model ducks [39], and finally *RoboRace*[3] - a high-speed big budget racing in specially designed race cars on full size circuits around the world. However, the most popular AV competition rooted firmly in the automotive industry is the international *Formula SAE Autonomous* competition[4] - with annual AV competitions taking place in different countries.

The competition comprises designing and building a fully autonomous vehicle from the ground up including a custom array of sensors and software to a base vehicle provided by the competition.

The driving tasks include: (a) straight-line acceleration; (b) a single lap of a track; (c) figure-of-eight (where the vehicle must handle an intersection; and (d) a 10-lap timed event, where the vehicle can learn from each lap in order to increase speed on the subsequent laps.

Each of the events involve following a course which is marked out by coloured traffic cones demarcating the circuit boundaries - with yellow and blue cones on left and right respectively, and orange cones to dictate other features (*e.g.,* start/finish lines, stopping areas etc).

The task for the AV's perception system is to accurately detect and correctly identify the location and colour of each cone. It must do so in the wide range of weather and lighting conditions that could occur at the racetrack. A mis-detection, or incorrect detection, could result in (a) a slow lap, (b) hitting a cone, which automatically earns a time penalty and makes a subsequent lap more challenging due to the misplaced or knocked-over cone, or (c) the vehicle veering off course resulting in failure to complete the event altogether.

It is therefore imperative to build a detailed understanding of how the AV's perception system will perform in all possible conditions that could occur at the racetrack, identifying and subsequently addressing the weaknesses within the ML model.

This case study demonstrates the potential of the use of the method proposed in this paper by testing, evaluating, and ultimately enhancing the performance of the ML object

[1] URL: https://www.formulapi.com/rules
[2] URL: https://iros2021.f1tenth.org/rules.html
[3] URL: https://roborace.com/
[4] URL: https://www.imeche.org/events/formula-student/team-information/fs-ai

detector used in an autonomous racing vehicle in a real-world setting on a racetrack.

## 4.3 The ML Model Under Test

The ML model used in this case study is the perception system of an autonomous racing car with input from a camera fitted on the vehicle. It is a custom trained YOLOv5 instance of "S" scale.

The YOLO models treat an object detection task as a multivariate regression problem. It outputs a sequence of detected instances with their classes and the coordinates of the bounding boxes after filtered by a confidence level and IOU (intersection over union). Being a regression task on those terms also means this model remains sensitive to changes in position within an image as well as bounding box size.

As shown in Figure 1, the architecture of YOLO neural network can be broken down into three sections. The *Backbone* is in charge of extracting feature maps of increasing contextual information and decreasing resolution from the input. The *Neck* then aggregates feature maps from different depths within the Backbone to retain as much of the contextual and pixel information as possible, resulting in three different levels of feature maps by default of different sizes. Finally, these aggregate feature maps are fed to the *Head*, which performs the multivariate inference and generates an output. In this step, several anchor combinations are used as per the default settings. Anchor sizes can however be auto generated to better adapt to the data used.

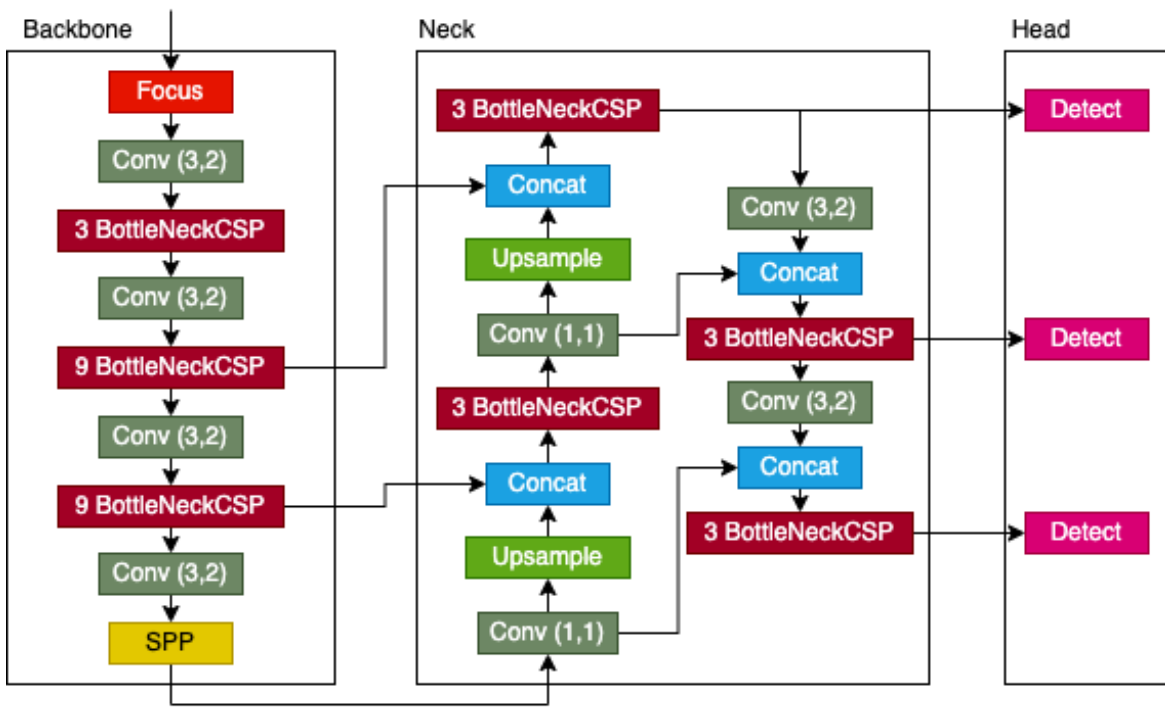

*Figure 1. Architecture of the ML Model*

The particular YOLO model used by the OBR Autonomous team as the image perception system is YOLOv5, which was originally pre-trained using the COCO dataset, but it is further fine-tuned or trained with custom data samples to specialise for the given task.

The training dataset contains 644 images, which were collected by the OBR Autonomous team through setting up cones around various scenarios with blurry scenes, confusing-colour objects in the background, dark lighting, slanted cones, and bright light but in a small ratio. The test dataset contains 100 images collected during the early stage of the project from a setup track on the campus with some dark images, objects in shadow, sun beam, etc.

## 4.4 Testing Process

The testing process started with an evaluation of the performance of the above ML model (referred to as model $M_0$ in the sequel) using three performance metrics: *precision*, *recall* and *mean average precision* (mAP); see Table 1 for their definitions.

### *4.4.1 Testing and Improving Model $M_0$*

The testing of model $M_0$ shows that the model's performance reached 93.01% on precision, 92.12% on recall and 90.13% on mAP@50, which is a fairly good performance.

*Table 1. Definitions of the Performance Metrics*

| **Metric** | **Definition** |
|---|---|
| Precision | The percentage of recognised cones that are correct with IoU = 0.5. |
| Recall | The percentage of cones that are recognised with IoU = 0.5 |
| mAP | The average precisions over all images, which the area under the precision-recall curve with IoU = 0.5 |

To further improve its performance, the errors made by the model were manually inspected and the problem is identified that many of the errors were in the situation when the picture was taken in a poor lighting condition and weather conditions.

Based on the analysis, we defined the following eight scenarios as suspected weak scenarios.

- *Bright*. The input image to the perception system is in a very bright lighting condition.
- *Dark*. The input image is in dark lighting condition.
- *Flare*. The input image is of flaring lights.
- *Rain*. The image is in a raining weather condition.
- *Fog*. The image is in a fog weather condition.
- *Water*. The image is in the situation when water is splashed on the camera lens.
- *Speed*. The input image is when the vehicle is moving fast.

These operation conditions were insufficiently represented in the training and testing datasets due to difficulty in the collection of such data and the expense in labelling.

To test the ML model on these scenarios, we developed a set of datamorphisms to transform normal images into images that have the corresponding features. They are implemented by using the library code developed by the open-source projects Automold[5] with some customisations to achieve required augmentation effect. Table 2 gives the details about how augmentations are implemented. Figure 2 shows some examples of the generated images using these image augmentations.

These datamorphisms are applied to the original test data that consists of 100 images and generated 700 mutant test cases. It is worth noting that, theoretically speaking, these datamorphisms preserves the semantics of the perception system in the sense that a cone in the original image should be recognised as the same cone in the augmented image. Therefore, the labels on the test cases should remain the same after augmenting the image. However, in reality, some of the cones in the original image may not be recognisable because, for example, completely disappeared behind thick fog. Therefore, for each mutant image, the labels were filtered manually. That is, if a label on a cone is not recognisable by a human tester, it was removed.

[5] URL: https://github.com/UjjwalSaxena/Automold--Road-Augmentation-Library

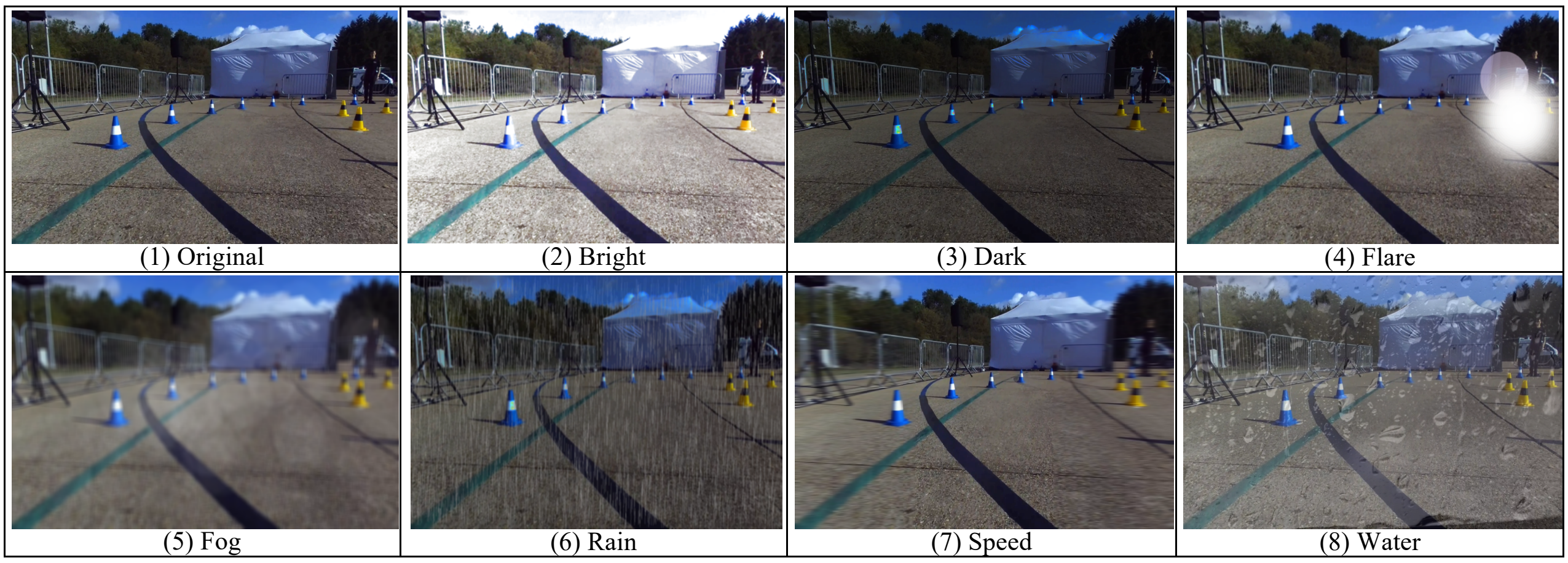

*Figure 2. Effects of Datamorphisms*

The ML model $M_0$ were tested on these filtered mutant test cases. The test data showed that the model's performance on such scenarios is less satisfactory that its precision, recall and mAP were 92.41%, 91.36%, 87.91%, respectively. Figure 3 shows $M_0$'s performance on each scenario.

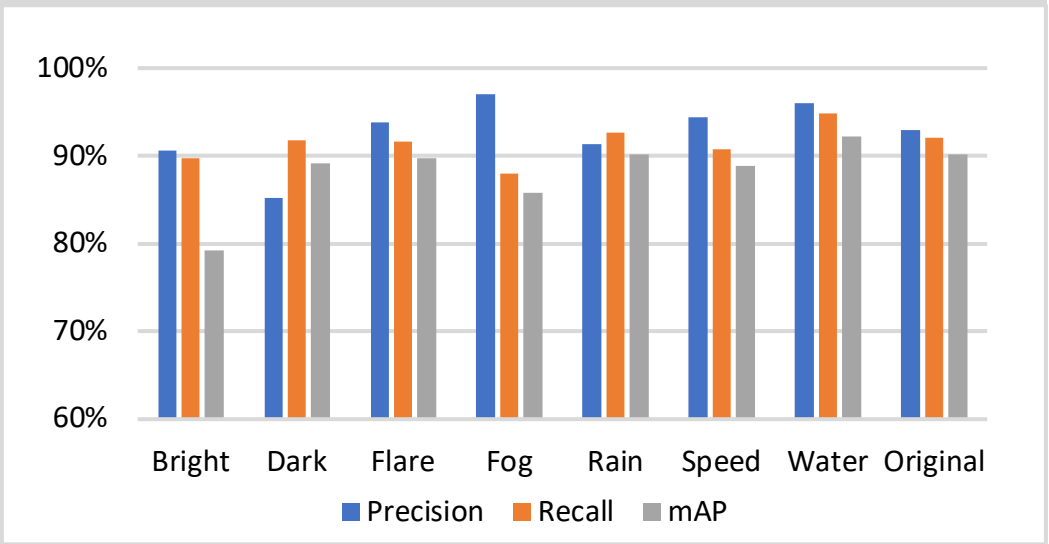

*Figure 3. Performance of $M_0$ on Different Scenarios (in mAP%)*

To improve the performances on these scenarios, the datamorphisms are applied to 10% of the $M_0$'s training data selected at random and selected another 10% original training data also at random to form a new training dataset. It is applied to model $M_0$ and obtained model $M_1$.

The model $M_1$ was then tested on these scenarios with the mutant test cases. The test data showed that $M_1$ improved overall performance by 1.62 percentage points on mAP and in most of the scenarios except Speed and Water; see Figure 5.

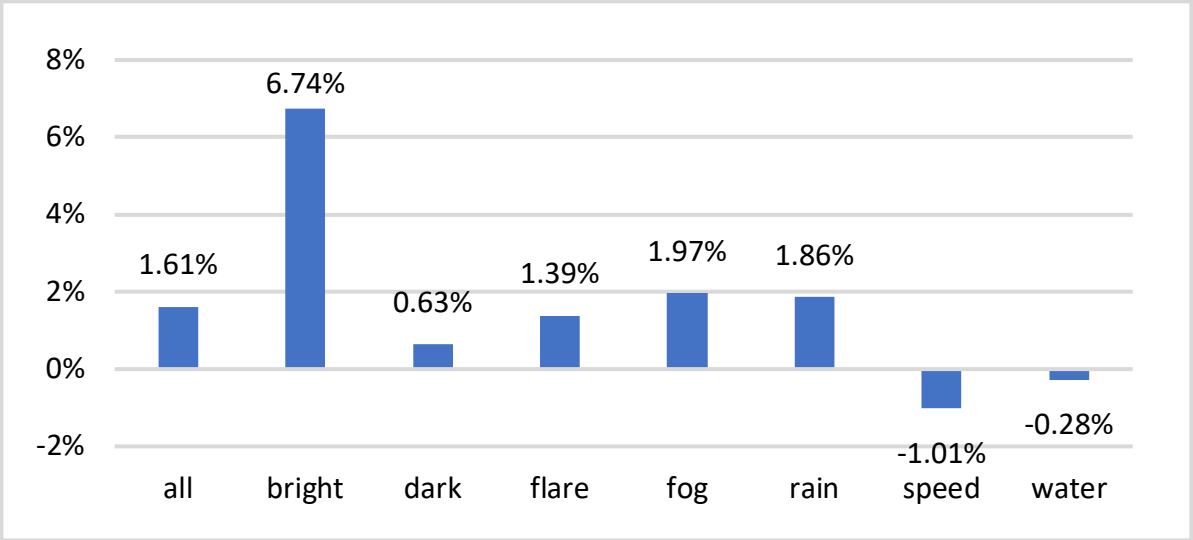

*Figure 5. $M_1$'s Improvement w.r.t. $M_0$ (in mAP%)*

*Table 2. Datamorphisms of the Test System*

| Operator | Specification | Implementation |
|---|---|---|
| Bright | Change the brightness of the image upwards | Set the bright coefficient = 0.9 |
| Dark | Change the brightness of the image downwards | Set the darkness coefficient = 0.4 |
| Flare | Add flare areas to the image | Add a flare layer |
| Fog | Add fog effect to the image | Set the fog coefficient = 0.4 |
| Rain | Add rain effect to the image | Add raining effect |
| Speed | Blur the image as if shot when camera is moving | Add a speedy effect |
| Water | Add water splash effect to the image | Add blurry effect and a layer of water drops |

### 4.4.2 *Testing and Improving Model $M_1$*

Although model $M_1$ improved the overall performance but its performance slightly decreased on the scenarios of Speed and Water. Further investigation of the reasons why no improvements was made on these scenarios. Manual inspections of the test cases on which the model fails indicated that the models did not detect orange-coloured cones so well as other coloured cones.

To confirm this assumption, $M_1$'s performances on detecting different types of cones were statistically analysed. The results are shown in Figure 4. It clearly indicates that $M_1$'s performance on detecting orange cones was weaker than detecting other types of cones.

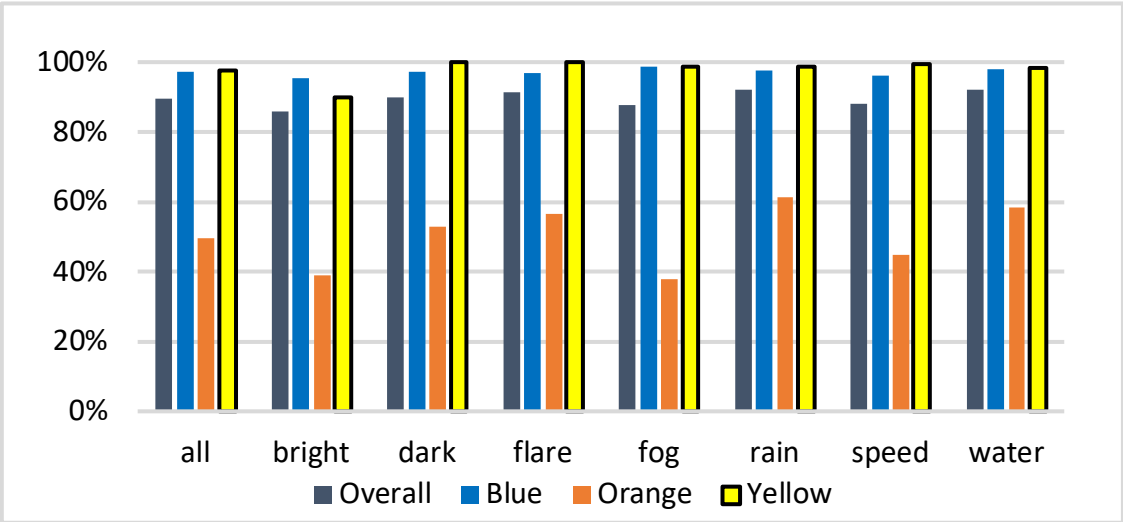

*Figure 4. $M_1$ Performances on Different Types of Cones (in mAP%)*

The reason why the model is weaker on detecting orange cones is that the training data contained significantly fewer orange cones than other types of cones; see Table 3.

*Table 3. Numbers of Cones in Test and Training Datasets*

| Dataset | Yellow | Blue | Orange | Total |
|---|---|---|---|---|
| Train | 2324 | 2349 | 65 | 4738 |
| Test | 77 | 347 | 96 | 520 |

In order to generate training data that contains sufficient orange cones, a datamorphism called *OrangeCone* was developed to transform blue cones in images into orange cones. It modifies all the blue pixels into orange in the detect box of blue cones.

A number versions of new model $M_{2.x}$ were build based on $M_1$ by further training with $M_1$ as the base. The OrangeCone datamorphism was applied to a subset of the images in the original training dataset selected at random to generate a set of synthetic data. The training used variable numbers of these synthetic data varying from 10% to 50% of the original training data plus 10% of the original training data also selected at random. Figure 6 shows the performances of the result models over $M_{2.x}$ when tested on 100 original test cases plus 700 mutant test cases with labels manually checked and corrected.

The data show that the performance peaked when the training data reaches 30% of the original training data. The performance of the model on detecting orange cones was significantly improved by 30% from model $M_1$ (50.05%) to $M_{2.3}$ (65.84%) at essentially zero cost.

While synthetic training data were used to improve the model, real data were collected by a different team through taking pictures of the orange cones. A total of 145 images were taken plus 10% of the original training data in the training of a new model based on $M_1$. The result model $M_{2.R}$ was tested in the same setting as other $M_2.x$ models. The results showed that $M_{2.3}$ outperforms $M_{2.R}$.

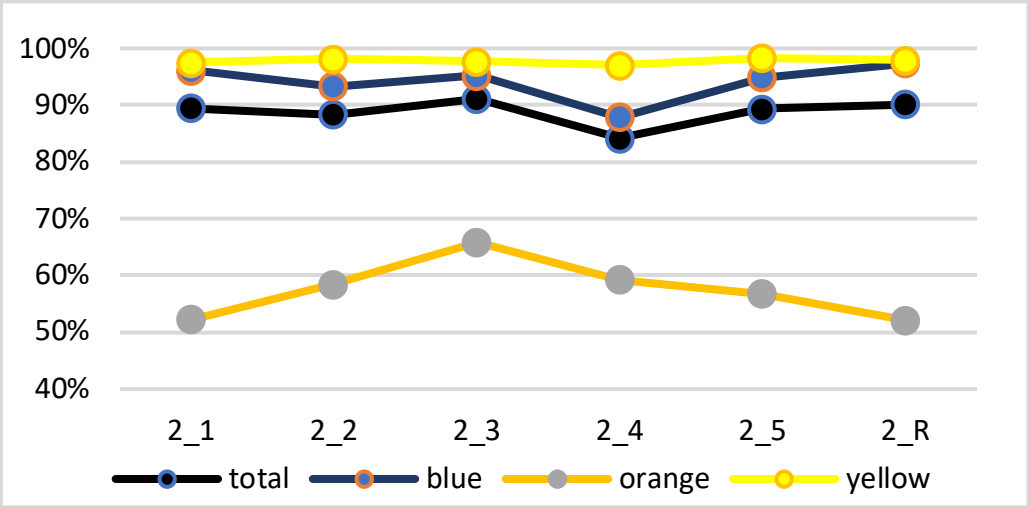

*Figure 6. Comparison of Different Versions $M_2$ (in mAP%)*

### 4.5 Discussion

In our case study, we observed a number of interesting phenomena that worth pointing out.

First, the case study demonstrated that the performance of DNN models can be improved through SBT following the iterative evolution process proposed in this paper. It is worth noting that ML models can never be perfect because of its inductive inference nature. In our case study, model $M_2$ is not perfect and there is still space for further improvement. Select a right scenario to improve performance is an important step of the test-evaluate-improve cycles. Assumptions made during the manual inspection of failed test cases may not lead to performance improvement. Statistical analysis of the assumption through SBT is necessary.

Second, our case study has also demonstrated that using synthetic data generated by applying augmentation to train machine learning models can achieve a performance as good as natural data. Development of datamorphisms to generate training data and test data is a cost-efficient and effective approach. Even if the datamorphism does not preserve the semantics completely, it can still significantly reduce the workload on labelling the data as show in our case study.

Third, it is widely believed that, the more training data the higher performance the model will achieve. However, in our case study, it is observed that the performance reaches the peak when the number of additional training data is at 30% of the size of original training data.

Finally, we applied a simple transfer learning technique to improve the ML model's performance on a specific scenario, i.e., using the existing model as the start point for training. A well-known phenomenon of this technique is "*catastrophic forgetting* (aka *catastrophic inference*)", which is the tendency for a model to compromise performance on previously learned data in favour of a higher performance in new data. It happens commonly in continual learning, reinforcement learning and transfer learning. In our case, the result model may demonstrate lower performance on other scenarios. Our approach to deal with this is to include a subset (10%) of the original training data in the re-training. This type of mitigation is called a rehearsal-based method in literature. Our experiment data shows that this is successful since the performances on other scenarios remain at the same level.

## 5 Conclusion

In this paper, we proposed a scenario-based functional testing methodology for improving ML models performances. The process emphasises on an iterative exploration testing of the model that each cycle consists of testing, diagnosing, treatment and testing again. The key factor of success in this process is how to improve a ML model's performance on a specific scenario without causing the side-effect of "forgetting" on other scenarios. Our solution is to apply a transfer learning technique with training dataset contains not only the data representing the treated scenario but also a subset of data from the original training dataset.

Our case study demonstrated that the approach is effective in the sense that DNN model's performance can be improved not only on the treated scenarios, but also the overall performance and side-effects can be prevented. It is also cost efficient that much less computation resources are required for preparing training and testing data and training the model in comparison with re-training the model from scratch.

The development of the perception system employed in the case study is still going on. The ML model $M_2$ still has a space to improve its performance. We are further analysing its weakness and improving it. The methodology will also be applied to test and improve other ML components of the autonomous racing car, which include a path planning system and a vehicle control system [40].

The case study is carried out by using the Morphy test automation tool. The identification of the weakness of the ML model in the case study largely relied on manual inspection of the erroneous test cases. It is supported by the Morphy's test case filtering facility, which enables erroneous test cases are collected and displayed easily. A test system including datamorphisms, metamorphisms and analysers are also implemented and in the Morphy, which enables repeated testing and evaluation automated. It is worth studying how the test system can be generalised and further supported by automated testing tools.

How to improve a ML model's performance is the heart of ML research problems. This paper proposed a scenario-based functional testing approach which identifies and then targets on weak scenarios in order to gain overall performance increases. The case study shows that employment of transfer learning is promising. It is worth further research. How to preserve the performances on other scenarios is an interesting and important problem for research of ML techniques.

## Acknowledgement

The work reported in this paper is partly funded by the Oxford Brookes University's 2020 Research Excellence Award. The authors are grateful to the OBR Autonomous team for their engagement in the project and providing support in the case study.